\documentclass[preprint,12pt]{elsarticle}
\usepackage{natbib}
\usepackage{amssymb}
\usepackage{amsmath}
\usepackage{graphicx}
\usepackage{algorithm}
\usepackage{algpseudocode}
\usepackage{array}   
\usepackage{booktabs} 
\usepackage{multirow}
\usepackage{bbm}
\usepackage{url}
\usepackage{microtype}
\usepackage{xcolor}
\biboptions{authoryear}

\journal{Expert Systems with Applications}

\begin{document}

\begin{frontmatter}

\title{Safety-Aware Evaluation of LLM-Generated Driver Intervention Messages through Multi-Task Risk Fusion}

\author[1,2]{Keito Inoshita\corref{cor1}}
\ead{inosita.2865@gmail.com}

\cortext[cor1]{Corresponding author.}

\affiliation[1]{organization={Faculty of Business and Commerce, Kansai University},
            addressline={3-3-35, Yamatecho}, 
            city={Suita},
            postcode={564-8680}, 
            state={Osaka},
            country={Japan}}
            
\affiliation[2]{organization={Data Science and AI Innovation Research Promotion Center, Shiga University},
            addressline={1-1-1 Baba},
            city={Hikone},
            postcode={522-8522},
            state={Shiga},
            country={Japan}}

\begin{abstract}
Existing driver intervention systems rely on auditory alerts and fixed templates, failing to leverage multi-task recognition outputs. General-purpose metrics such as BLEU and BERTScore cannot capture intervention-specific quality dimensions including risk-urgency alignment, cognitive load, and driver acceptability. In this paper, we propose the Driver Safety-Aware Intervention Score (DSAIS), a domain-specific metric evaluating five dimensions through a hybrid architecture combining lightweight rule-based computation with LLM Judge evaluation, together with an end-to-end framework integrating four-task recognition outputs into an LLM through risk fusion, state history management, and dynamic prompt construction. Experiments on the AIDE dataset with five models and seven conditions demonstrate that DSAIS achieves ICC 0.798--0.840 across three architecturally distinct judges and Cohen's $d > 1.5$ across all control conditions. Multi-dimensional sub-score analysis quantifies the contextual adaptability gap between rule-based and LLM-based systems, revealing that multi-task integration improves contextual relevance by 9.1\% over rule-based baselines. Ablation experiments demonstrate that each framework component contributes to contextual relevance, with sub-score decomposition revealing gains that aggregate scoring masks. Driver emotion recognition is identified as the most critical upstream factor, and compact local LLMs (7B--9B parameters) achieve quality superior to API-based models, providing practical design guidelines for in-vehicle deployment.
\end{abstract}




\end{frontmatter}

\section{Introduction}\label{sec:intro}

Road traffic crashes remain a leading cause of death and injury worldwide. According to the National Highway Traffic Safety Administration (NHTSA), distracted driving was involved in approximately 3,021 fatal crashes and 3,275 fatalities in 2023, accounting for roughly 8\% of all traffic deaths~\citep{1}. Drowsy driving is officially attributed to 1.5\% of fatal crashes; however, the AAA Foundation for Traffic Safety estimates that, when underreporting is taken into account, drowsy driving was involved in 17.6\% of fatal crashes~\citep{2}. These statistics underscore the critical importance of timely and appropriate driver intervention.

Modern driver monitoring systems have advanced considerably, evolving from single-task detectors to multi-task architectures capable of simultaneously recognizing traffic context, vehicle dynamics, driver emotion, and driver behavior~\citep{3}. However, even when these systems correctly detect hazardous states, interventions are predominantly implemented as attentional warnings, which may indiscriminately treat context-dependent gaze allocation and driving-related behaviors as dangerous~\citep{4}. For instance, in a compound hazardous state where a driver exhibits signs of fatigue while making a phone call and changing lanes in heavy traffic, conventional systems merely display a generic message such as ``Please focus on driving,'' without addressing any of these interacting factors. Large Language Models (LLMs) offer a promising pathway toward such context-adaptive intervention~\citep{20}. Their ability to interpret structured inputs, reason about relationships among multiple factors, and generate fluent natural language makes them a natural candidate for bridging the gap between detection and intervention. However, the following challenges remain unresolved in applying LLMs to driver intervention. 

The first challenge is the lack of appropriate means to evaluate the quality of intervention messages generated by LLMs. Existing Natural Language Generation (NLG) evaluation metrics such as BLEU~\citep{6}, ROUGE~\citep{7}, and BERTScore~\citep{8} are limited to measuring superficial or semantic similarity to reference texts. These metrics cannot assess whether a message appropriately matches the urgency of the situation, whether its length is compatible with in-vehicle reading constraints, or whether its tone is likely to be accepted by a stressed driver. Without a domain-specific evaluation metric, it is difficult to systematically analyze LLM intervention capabilities or to compare model characteristics across multiple quality dimensions. The second challenge is the absence of an established framework for integrating multi-task recognition outputs into an LLM. Prior work has explored LLMs for behavioral science interventions~\citep{9} and clinical text generation~\citep{10}, but the unique requirement that the consumer of the generated text is a cognitively loaded driver operating a vehicle has not been systematically addressed. A structured framework is needed that preserves each task's contribution while computing an integrated risk assessment and generating messages tailored to the driver's state. The third challenge is that the impact of upstream recognition accuracy on downstream intervention quality has not been quantitatively understood. Identifying which task's misclassification causes the most severe degradation provides direct guidance for computational resource allocation in driver monitoring system design~\citep{11}. In addition, as on-device deployment becomes increasingly important for privacy and latency reasons~\citep{12}, whether compact open-source LLMs can match the performance of large-scale API-based models on this task is a practically important question.

To address these challenges, in this study, we propose an end-to-end framework that bridges multi-task driver state recognition and LLM-based natural language intervention, together with the Driver Safety-Aware Intervention Score (DSAIS), a domain-specific metric for multi-dimensional quantitative evaluation of intervention quality. DSAIS independently evaluates five quality dimensions (risk-tone alignment, contextual relevance, conciseness, cognitive load, and driver acceptability) and is designed with lightweight rule-based computation as the primary method, supplemented by LLM Judge evaluation, thereby ensuring both continuous quality monitoring in in-vehicle environments and reproducibility of evaluation. By conducting systematic experiments using DSAIS on the proposed framework, we derive intervention-specific insights that were unattainable with conventional general-purpose NLG metrics, specifically, which task's recognition accuracy most critically affects intervention quality, and which model excels along which quality dimension. The main contributions of this paper are summarized as follows:

\begin{enumerate}[i)]
\item A domain-specific composite metric, DSAIS, is newly designed to evaluate five intervention quality dimensions through a hybrid architecture combining rule-based computation with LLM Judge evaluation. Its reliability and discriminative power are validated through multi-judge agreement analysis and control condition experiments.
\item An end-to-end intervention framework is newly constructed, in which four-task recognition outputs are integrated into an LLM through risk fusion, state history management, and dynamic prompt construction. The framework achieves the highest contextual relevance ($S_\text{context}$) with Qwen2.5-7B, improving by 2.7\% over single-task and 9.1\% over rule-based baselines, with the smallest output variance among all methods.
\item Systematic misclassification experiments across five models and seven conditions, together with comprehensive ablation and sensitivity analyses, are conducted to reveal that driver emotion recognition is the most critical upstream factor for intervention quality. Compact local LLMs (7B--9B) are demonstrated to achieve quality superior to API-based models, providing practical design guidelines for in-vehicle deployment. 
\end{enumerate}

The rest of this paper is organized as follows. Section~\ref{sec:related} reviews related work on driver state monitoring, LLM-based intervention systems, and NLG evaluation metrics. Section~\ref{sec:method} introduces the proposed framework and the DSAIS metric. Section~\ref{sec:experiments} presents the experimental design, results, and ablation and sensitivity analyses. Section~\ref{sec:discussion} discusses the insights, design guidelines, and limitations. Finally, we conclude this study in Section~\ref{sec:conclusion}.
 
\section{Related Work}\label{sec:related}

\subsection{Driver State Monitoring Systems}

Research on driver state monitoring has rapidly evolved from single-task detectors to integrated multi-view, multi-modal, multi-task recognition. Early studies focused on individual tasks such as drowsiness detection~\citep{13} and head pose estimation~\citep{14}, pursuing accuracy improvements for each task independently. Subsequently, Drive\&Act~\citep{15} provided a multi-modal dataset for fine-grained recognition of driving behaviors, establishing the direction for behavior category granularity and benchmarking. DeepTake~\citep{16} integrated vehicle, physiological, and subjective data to predict takeover intent, timing, and quality, demonstrating the effectiveness of multi-modal feature integration for driving assistance.

In recent years, comprehensive datasets and benchmarks integrating in-cabin and external sensing have been developed. AIDE~\citep{3} designed four tasks---Traffic Context Recognition (TCR), Vehicle Context Recognition (VCR), Driver Emotion Recognition (DER), and Driver Behavior Recognition (DBR)---in a multi-view, multi-modal setting and presented benchmark experiments including fusion strategies. DANet~\citep{17} demonstrated an efficient multi-task learning approach for driver attention monitoring. DriverMVT~\citep{18} improved monitoring reproducibility through a dataset combining in-cabin video with vehicle telemetry. manD~\citep{19} provided in-cabin monitoring data under complex automation scenarios, presenting data configurations under realistic driving conditions.

These advances have enriched upstream recognition outputs, making it technically feasible to simultaneously capture driver emotion, behavior, traffic context, and vehicle state. A systematic review of driver situation awareness for regaining control from automated vehicles~\citep{49} further confirms that multi-dimensional driver state understanding is essential for effective intervention design. However, the downstream side, converting recognition results into driver interventions, still relies on fixed templates and simple auditory alerts, leaving a fundamental gap where the achievements of multi-task recognition are not leveraged to improve intervention quality. This study addresses this gap by proposing a framework for intervention message generation that integrates four-task recognition outputs, together with a metric for multi-dimensional evaluation of intervention quality.

\subsection{LLM-Based Intervention and Warning Systems}

Research on in-vehicle applications of LLMs is expanding rapidly and can be organized into three streams. The first stream concerns personalization and modality planning of driver warnings and interventions. \citet{20} proposed an LLM-based personalized warning agent comprising four modules (memory, perception, control, and action), demonstrating the direction of dynamically optimizing warnings based on driver-specific context rather than static user settings. The same group also proposed a system that plans multi-modal warnings from traffic environments and driver profiles~\citep{21}, raising the point that modality selection, not just content, affects intervention quality.

The second stream concerns architectures for integrating LLMs into vehicle systems. \citet{22} positioned an LLM as a vehicle co-pilot, presenting an early framework encompassing dialogue, planning, and decision support. ConnectGPT~\citep{23} constructed a pipeline for generating standardized C-ITS messages from traffic situations, ensuring controllability of free-text generation through mapping to standardized messages. \citet{24} introduced a hierarchy of proactive levels for in-vehicle conversational assistants and presented LLM operation strategies (Rewrite+ReAct+Reflect) with human evaluation. Talk2Drive~\citep{46} demonstrated personalized autonomous driving through LLM-based natural language commands, reducing takeover rates by 75.9\% through memory-augmented personalization in field experiments.

The third stream concerns side effects of LLM-equipped systems on driving. \citet{25} experimentally demonstrated the impact of voice assistant conversational style on cognitive driver distraction, revealing the design trade-off that natural conversation is not necessarily safe. \citet{26} reported an attempt to mitigate passive fatigue in conditional automated driving through LLM-based conversation, and a subsequent full study~\citep{47} confirmed that LLM-based conversational agents effectively maintain driver vigilance under passive fatigue. \citet{45} proposed context-aware advisory warnings (CAWA) that adapt warning modalities to non-driving-related task contexts, demonstrating significant improvements in safe takeover behavior and situation awareness. Although predating LLM-based approaches, \citet{27} demonstrated the effectiveness of monitoring requests for restoring driver attention prior to takeover requests, suggesting that interventions should be designed as staged, contextual sequences rather than single messages. \citet{50} proposed a hierarchical risk taxonomy for safety-critical LLM-based driving assistants, identifying 129 failure mode categories across technical, legal, social, and ethical dimensions.

These prior studies have advanced in-vehicle LLM applications from the perspectives of personalization, vehicle integration, and side effects, respectively. However, no existing framework simultaneously addresses the integrated utilization of multi-task recognition outputs, analysis of how upstream recognition accuracy affects downstream intervention quality, and domain-specific quality evaluation of driver interventions. The proposed framework is positioned to cut across these streams by integrating risk fusion, state history management, and dynamic prompt construction.

\subsection{Evaluation Metrics for Generated Text}

Evaluation of NLG has progressively evolved from classical metrics based on reference-text agreement, through learned metrics and consistency evaluation, to LLM-as-a-Judge approaches. BLEU~\citep{6} is a representative reference-based metric grounded in n-gram overlap, and together with ROUGE~\citep{7}, has formed the foundation of NLG evaluation. However, these metrics have limitations when generated texts are diverse and no single correct reference exists. BERTScore~\citep{8} extended evaluation to semantic similarity via contextual embeddings, and BLEURT~\citep{28} pursued improved correlation with human judgments as a learned metric. All of these operate within a framework of measuring similarity to reference texts and cannot assess whether generated text is logically consistent with the input situation or satisfies domain-specific requirements.

To address this limitation, SummaC~\citep{29} proposed a framework for detecting inconsistencies between generated summaries and source inputs using natural language inference. G-Eval~\citep{30} employed powerful LLMs such as GPT-4 as evaluators, improving alignment with human evaluation through explicit evaluation criteria and Chain-of-Thought reasoning. MT-Bench~\citep{31} demonstrated the utility of LLM-as-a-Judge while also analyzing evaluator-specific issues such as position bias and verbosity bias, and discussing design countermeasures. A comprehensive survey by \citet{43} organized LLM-as-a-Judge approaches along three dimensions (what, how, and benchmarking), identifying systematic biases and vulnerabilities. \citet{44} further quantified these biases, demonstrating that verbosity bias and fallacy-oversight bias remain significant challenges for LLM-based evaluation. Prometheus 2~\citep{42} addressed these issues by training open-source evaluation models on rubric-based criteria, achieving correlation with human judgments competitive with proprietary models. SelfCheckGPT~\citep{48} proposed zero-resource hallucination detection through multi-sample consistency checking, providing a complementary approach to reference-free evaluation. As an evaluation dimension specific to the driving context, cognitive load measurement is relevant. NASA-TLX~\citep{32} is one of the most widely used subjective workload assessment scales and serves as a primary metric for human-centered evaluation of the cognitive burden that intervention messages and in-vehicle HMIs impose on drivers.

Considering the evolution of these evaluation metrics, no single approach is sufficient on its own for evaluating driver intervention messages. A domain-specific metric that integrates these quality dimensions is needed. The DSAIS proposed in this study independently evaluates five dimensions, and comprehensively addresses the limitations of prior work by combining reproducible rule-based computation with the contextual understanding of an LLM Judge.

\section{Multi-Task Recognition-Integrated LLM Intervention Framework}\label{sec:method}

\subsection{Framework Overview}\label{sec:overview}

\begin{figure}[t]
\centering
\makebox[\textwidth][c]{\includegraphics[width=1.3\textwidth]{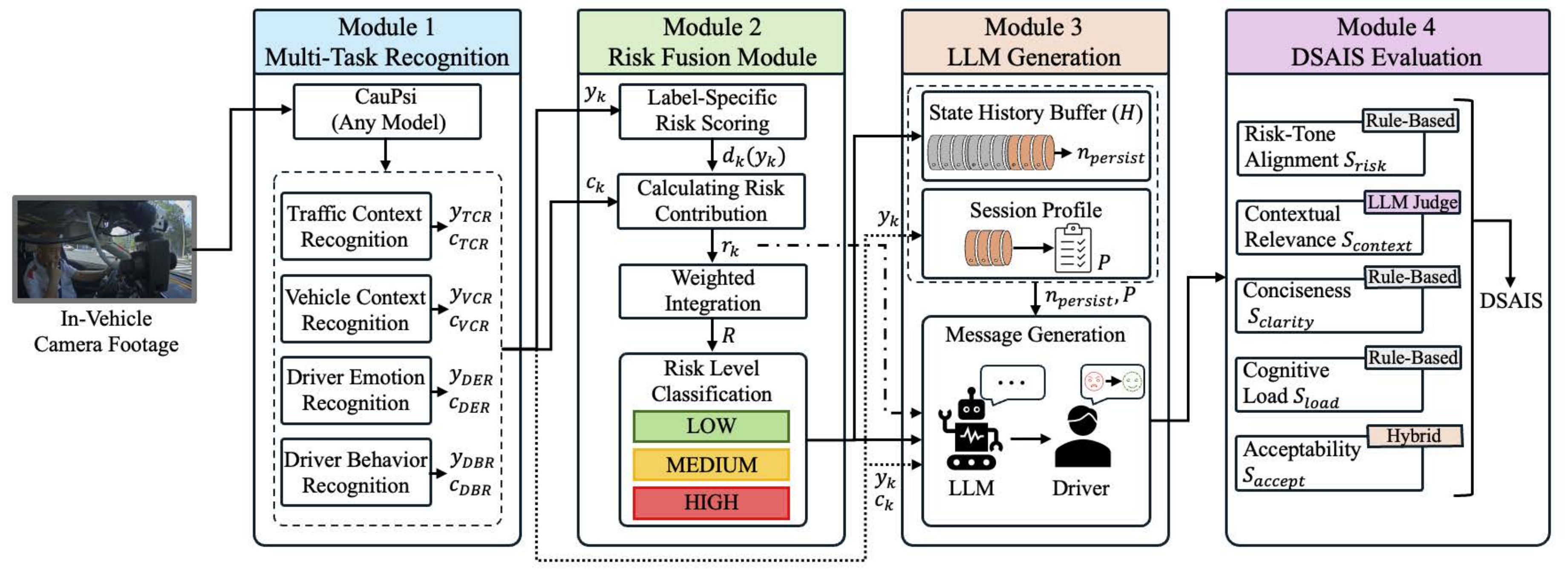}}
\caption{Overall architecture of the proposed framework.}
\label{fig:architecture}
\end{figure}

Figure~\ref{fig:architecture} illustrates the overall architecture of the proposed framework. The framework comprises four main modules and handles the entire pipeline from upstream recognition results to downstream intervention message evaluation. The first module, the multi-task recognition engine CauPsi~\citep{33}, simultaneously infers four tasks (TCR, VCR, DER, and DBR) from in-vehicle camera footage and outputs the predicted label and confidence score for each task. CauPsi represents the highest-performing model designed for in-vehicle deployment at the time of writing; its details are described in a separate paper, and this study uses its inference outputs as input to the subsequent modules.

The second module, the risk fusion module, receives the four-task inference results, multiplies each label's predefined risk score by its detection confidence, and computes an integrated risk score $R$ through a weighted linear combination. $R$ is discretized into three risk levels (LOW, MEDIUM, and HIGH) based on thresholds. In addition, this module manages state history through a ring buffer retaining the most recent $N$ frames of recognition results, and constructs a session profile recording the cumulative frequency of hazardous behaviors within the session.

The third module, the LLM generation module, receives the integrated risk score $R$, risk level, state history, and session profile as a structured prompt, from which the LLM generates a JSON-formatted output containing the intervention message, tone, and risk level assessment. The prompt is designed to dynamically reflect the driver's current state and past behavioral patterns, enabling the generation of personalized intervention messages.

The fourth module, the DSAIS evaluation module, provides multi-dimensional quantitative evaluation of the generated message quality. It comprises five sub-scores, risk-tone alignment ($S_\text{risk}$), contextual relevance ($S_\text{context}$), conciseness ($S_\text{clarity}$), cognitive load ($S_\text{load}$), and driver acceptability ($S_\text{accept}$), and employs a hybrid design combining rule-based computation with LLM Judge evaluation to balance reproducibility and contextual understanding.

The input to the framework is defined as the inference output of a multi-task recognition model:
\begin{equation}\label{eq:input}
X = \left(y_\text{TCR}, c_\text{TCR}\right),\ \left(y_\text{VCR}, c_\text{VCR}\right),\ \left(y_\text{DER}, c_\text{DER}\right),\ \left(y_\text{DBR}, c_\text{DBR}\right)
\end{equation}
where $y_k$ is the predicted label for task $k$ and $c_k \in [0,1]$ is its confidence score. To ensure framework generality, two types of input conditions are distinguished: the Ground-Truth (GT) condition $X_\text{gt}$, which uses correct labels, and the Prediction condition $X_\text{pred}$, which uses actual inference results from the recognition model. The $X_\text{gt}$ condition measures the theoretical upper bound of LLM interpretation capability, while the $X_\text{pred}$ condition evaluates overall performance under real-world operating conditions. Comparison of these two conditions enables systematic analysis of the impact of upstream recognition accuracy on downstream intervention quality. The design of specific experimental conditions based on these two conditions is described in Section~\ref{sec:experimental_conditions}.

The objective of this framework is to generate a natural language intervention message $m$ that simultaneously satisfies the following four requirements: i)~appropriate message content based on the compound driver state across four tasks, ii)~urgency calibrated to the integrated risk level, iii)~sufficient conciseness for in-vehicle consumption, and iv)~expression likely to be accepted by the driver. These requirements directly correspond to the five sub-scores of DSAIS proposed in Section~\ref{sec:dsais}.

\subsection{Risk Fusion Module}\label{sec:risk_fusion}

The risk fusion module converts the discrete inference results of four tasks into a single continuous value, the integrated risk score $R$, and performs risk level classification and state history management.

\subsubsection{Label-Specific Risk Scores}

Each label within each task carries a different risk implication. For example, within DER, Happiness poses negligible risk, whereas Anger carries high risk due to its association with impulsive driving behavior. A label-specific risk score $d_k(y) \in [0,1]$ is defined as a lookup table for each task $k$ and label $y$, as listed in Table~\ref{tab:risk_scores}.

\begin{table}[t]
\centering
\caption{Label-specific risk scores $d_k(y)$ for four tasks. Each value is defined in the range 0 (safe) to 1 (maximum risk).}
\label{tab:risk_scores}
\small
\begin{tabular}{llcl}
\toprule
Task & Label & $d_k(y)$ & Rationale \\
\midrule
\multirow{5}{*}{DER} & Happiness & 0.0 & Stable emotional state \\
 & Peace & 0.1 & Calm state \\
 & Anxiety & 0.6 & Risk of distraction/hyper-tension \\
 & Weariness & 0.8 & Reduced reaction due to fatigue \\
 & Anger & 0.9 & Risk of impulsive maneuvers \\
\midrule
\multirow{7}{*}{DBR} & Normal Driving & 0.0 & Normal operation \\
 & Talking & 0.2 & Mild distraction \\
 & Body Movement & 0.3 & Brief distraction \\
 & Looking Around & 0.5 & Lack of forward gaze \\
 & Smoking & 0.6 & Hand/gaze distraction \\
 & Making Phone & 0.8 & Severe distraction \\
 & Dozing Off & 1.0 & Maximum risk \\
\midrule
\multirow{3}{*}{TCR} & Smooth Traffic & 0.1 & Low-risk environment \\
 & Waiting & 0.2 & Stationary/low speed \\
 & Traffic Jam & 0.6 & Rear-end/hard braking risk \\
\midrule
\multirow{5}{*}{VCR} & Parking & 0.0 & Stationary \\
 & Forward Moving & 0.2 & Normal driving \\
 & Turning & 0.4 & Requires focused maneuvering \\
 & Changing Lane & 0.6 & Combined judgment/maneuvering load \\
 & Backward Moving & 0.7 & Restricted visibility/reversing \\
\bottomrule
\end{tabular}
\end{table}

These risk scores were calibrated based on traffic safety statistics and prior research. For DER, the AAA Foundation estimates that drowsy driving was involved in an estimated 17.6\% of fatal crashes~\citep{2} informed the assignment of high risk scores to Weariness (0.8) and Anger (0.9). For DBR, NHTSA statistics reporting that distracted driving accounts for approximately 8\% of fatal crashes~\citep{1} motivated high values for Making Phone (0.8) and Dozing Off (1.0). Risk scores for TCR and VCR were set in a graded manner based on driving operational load and environmental risk considerations.

\subsubsection{Integrated Risk Score and Risk Level Classification}

The risk contribution $r_k$ for each task is defined as the product of the label-specific risk score $d_k(y_k)$ and the detection confidence $c_k$:
\begin{equation}\label{eq:task_risk}
r_k = d_k(y_k) \cdot c_k
\end{equation}
This formulation naturally attenuates the risk contribution from low-confidence predictions, reflecting the uncertainty of the upstream recognition model. The integrated risk score $R$ is computed as a weighted linear combination across all four tasks. Let $\mathcal{K} = \{\text{TCR, VCR, DER, DBR}\}$:
\begin{equation}\label{eq:integrated_risk}
R = \sum_{k \in \mathcal{K}} w_k \cdot r_k, \quad \sum_k w_k = 1
\end{equation}
Task weights are set as shown in Table~\ref{tab:task_weights}. The highest weight (0.35) is assigned to DER, reflecting both the widely reported role of fatigue as a critical crash risk factor and prior research demonstrating that driving anger and aggression are significantly associated with hazardous driving and crash involvement~\citep{34}.

\begin{table}[t]
\centering
\caption{Task weight settings.}
\label{tab:task_weights}
\small
\begin{tabular}{lccp{5.5cm}}
\toprule
Task & $w_k$ & Target & Rationale \\
\midrule
TCR & 0.20 & Traffic context & Environmental factor; not directly controllable but contributes to situational awareness \\
VCR & 0.20 & Vehicle state & Driving state provides maneuvering load context \\
DER & 0.35 & Emotional state & Involved in an estimated 17.6\% of fatal crashes \\
DBR & 0.25 & Behavioral state & Approximately 8\% of fatal crashes \\
\bottomrule
\end{tabular}
\end{table}

Risk level classification based on $R$ is performed using the following thresholds:
\begin{equation}\label{eq:risk_level}
\text{Level}(R) = \begin{cases}
\text{LOW} & R < \theta_L \\
\text{MEDIUM} & \theta_L \leq R < \theta_H \\
\text{HIGH} & R \geq \theta_H
\end{cases}
\end{equation}
In this study, $\theta_L = 0.3$ and $\theta_H = 0.6$ are adopted. The three-level discretization is motivated by three considerations: i)~excessively fine-grained levels impose unnecessary cognitive load on the driver, ii)~the three levels naturally correspond to three tones (gentle, warning, and urgent), and iii)~LLMs can reliably interpret three-level instructions.

\subsubsection{State History Management and Session Profile}\label{sec:history}

Single-frame state recognition alone cannot distinguish between transient false detections and persistent hazardous states. To address this, the module manages two types of temporal context information.

First, the most recent $N$ recognition outputs (default $N = 5$) are retained in a ring buffer:
\begin{equation}\label{eq:history}
H = \{X_{t-N+1}, \ldots, X_t\}
\end{equation}
From this buffer, the persistent risk count $n_\text{persist}$, representing the number of frames classified as MEDIUM or higher risk among the most recent $N$ frames, is computed:
\begin{equation}\label{eq:npersist}
n_\text{persist} = |\{i \in H \mid \text{Level}(R_i) \in \{\text{MEDIUM}, \text{HIGH}\}\}|
\end{equation}
A larger $n_\text{persist}$ indicates a persistent hazardous state, and the LLM should generate a stronger intervention message. The computation of $n_\text{persist}$ includes the current frame $X_t$ itself.

Second, a session profile $P$ manages the cumulative frequency of hazardous behaviors and emotions within the session:
\begin{equation}\label{eq:profile}
P = \{f_\text{DER}(l),\ f_\text{DBR}(l) \quad \forall l \in \text{Labels}\}
\end{equation}
where $f_k(l)$ denotes the cumulative count of label $l$ detected within the current session. In the implementation, the drowsiness-related counter aggregates both Weariness and Anxiety occurrences, as anxiety-induced hyper-arousal and fatigue-induced hypo-arousal both impair sustained attention and are associated with elevated crash risk in the traffic safety literature. The session profile enables the LLM to generate escalating interventions for recurrent risk behaviors. For example, when drowsiness-related states are detected multiple times for the same driver, the session profile serves as grounds for generating a stronger rest suggestion rather than a mere alert.

\subsection{LLM Generation Module}\label{sec:generation}

The LLM generation module receives the output of the risk fusion module as a structured prompt and generates natural language intervention messages. The prompt follows a two-tier structure consisting of a system prompt and a user prompt. The system prompt defines the LLM's role and output constraints, while the user prompt dynamically injects per-frame driver state information.

The main constraints of the system prompt are the following six items: i)~output language restricted to English, ii)~message length limited to 15 words or fewer, iii)~suggestive style preferred over imperative, iv)~urgency calibrated to the risk level, v)~personalization considering session history, and vi)~output format restricted to JSON. The LLM output includes the message body, a self-assessed tone (gentle/warning/urgent), and a risk level assessment in JSON format. The self-assessed tone field is designed for analyzing how the generation LLM perceives its own message, enabling independent verification of consistency between message content and meta-information. Accordingly, this tone field is not used in the computation of $S_\text{risk}$.

The user prompt is populated with the following structured information: predicted labels, confidence scores, and risk contributions $r_k$ for four tasks; the integrated risk score $R$ and its risk level; the persistent risk count $n_\text{persist}$; and key items from the session profile $P$ (e.g., drowsiness detection count, phone use count, anger detection count). The full prompt text is provided in Appendix~\ref{app:prompts}.

\subsection{Driver Safety-Aware Intervention Score}\label{sec:dsais}

DSAIS is a composite metric for multi-dimensional quantitative evaluation of generated message quality, consisting of five sub-scores. The metric is designed with continuous quality monitoring on in-vehicle systems in mind. As a fundamental design principle, dependence on LLM Judges is minimized, and lightweight, deterministic rule-based and linguistic analysis-based computation is adopted wherever possible. This ensures feasibility of execution in in-vehicle environments without reliance on external APIs, while guaranteeing evaluation reproducibility and transparency. Table~\ref{tab:dsais_overview} provides an overview of each sub-score. The definitions of individual sub-scores are described in the table.

\begin{table}[t]
\centering
\caption{Overview of the five DSAIS sub-scores.}
\label{tab:dsais_overview}
\small
\makebox[\textwidth][c]{%
\begin{tabular}{lll}
\toprule
Sub-score & Target & Computation \\
\midrule
$S_\text{risk}$ & Risk-tone alignment & Judge tone classification + continuous distance \\
$S_\text{context}$ & Contextual relevance & GPT-5 Judge 1--5 scale \\
$S_\text{clarity}$ & Conciseness & Rule-based (word count) \\
$S_\text{load}$ & Cognitive load & Rule-based (information elements, spaCy) \\
$S_\text{accept}$ & Driver acceptability & Hybrid (rule + Judge) \\
\bottomrule
\end{tabular}}
\end{table}

\subsubsection{Risk-Tone Alignment Score}\label{sec:s_risk}

This sub-score measures whether the tone of the generated message is aligned with the integrated risk score $R$. To ensure objectivity, tone classification is performed not by the message generation LLM but by an external GPT-5 Judge. The GPT-5 Judge receives only the message body as input (without $R$, risk level, or other context information) and classifies the tone into three categories: gentle/warning/urgent. The classification result is mapped to $t \in \{0, 0.5, 1.0\}$, and the continuous distance from $R$ is computed:
\begin{equation}\label{eq:s_risk}
S_\text{risk} = \max\left(0,\ \min\left(1,\ 1 - |t - R|\right)\right)
\end{equation}
For example, when $R = 0.55$ and the Judge classifies the tone as warning ($t = 0.5$), $S_\text{risk} = 1 - |0.5 - 0.55| = 0.95$. Conversely, when $R = 0.55$ and the Judge classifies the tone as gentle ($t = 0.0$), $S_\text{risk} = 0.45$, imposing a substantial penalty for inappropriate tone selection. When the LLM fails to produce valid JSON, $S_\text{risk} = 0.0$ is assigned as a structural constraint violation.

\subsubsection{Contextual Relevance Score}

This sub-score measures the logical consistency between the four-task recognition results and the message. The GPT-5 Judge is presented with the pair of four-task predicted labels and the message, and evaluates ``Is this message appropriate for this driving situation?'' on a 1--5 integer scale, normalized to $[0,1]$:
\begin{equation}\label{eq:s_context}
S_\text{context} = \frac{\text{judge\_score} - 1}{4}
\end{equation}
$S_\text{context} \in [0, 1]$, with higher values indicating messages that are more logically consistent with the situation. Judge API parameters are fixed at default values to ensure evaluation reproducibility. The full Judge prompt is provided in Appendix~\ref{app:judge_prompts}.

\subsubsection{Conciseness Score}

This sub-score measures whether the message can be immediately understood under the limited cognitive resources available during driving, using a rule-based score based on word count $W$ (space-delimited). Prior research has shown that the information complexity of in-vehicle warning messages affects driver cognitive load and response, with visually simple and concise warnings being advantageous~\citep{35}. Furthermore, on-board messages during driving need to be comprehended within a short time, and warning messages have been reported to be understood faster than advisory messages~\citep{36}. Based on these findings, this study sets the optimal word count at $W_\text{opt} = 8$ and the upper limit at $W_\text{max} = 15$:
\begin{equation}\label{eq:s_clarity}
S_\text{clarity} = \max\left(0,\ 1 - \frac{\max(0,\ W - W_\text{opt})}{W_\text{max} - W_\text{opt}}\right)
\end{equation}
$S_\text{clarity} = 1.0$ when $W \leq W_\text{opt}$ and $S_\text{clarity} = 0.0$ when $W \geq W_\text{max}$. This linear penalty assigns the highest score to concise messages of 8 words or fewer and fully penalizes messages exceeding 15 words. $W_\text{max} = 15$ is consistent with the system prompt constraint. Syntactic complexity is not included in $S_\text{clarity}$ because its discriminative power is extremely low for short texts of 15 words or fewer.

\subsubsection{Cognitive Load Score}

This sub-score measures the information density contained in the message and evaluates whether it imposes excessive cognitive load. Prior research has demonstrated that the information complexity of in-vehicle warning messages affects driver cognitive load and response, and that information-rich warnings can increase resource consumption and annoyance, whereas concise and visually simple warnings are advantageous~\citep{37}. On-board messages during driving must be read and understood within a short time~\citep{38}. Based on these findings, the number of information elements $I$ (occurrences of numerals, named entities, and deictic expressions) is extracted using spaCy~\citep{40} NER and POS tagging, with $I_\text{max} = 3$ set as the design threshold:
\begin{equation}\label{eq:s_load}
S_\text{load} = \max\left(0,\ 1 - \frac{I}{I_\text{max}}\right)
\end{equation}
Message length (word count) is already evaluated by $S_\text{clarity}$; therefore, $S_\text{load}$ excludes word count and focuses exclusively on information density, preventing double counting between the two scores.

\subsubsection{Driver Acceptability Score}

This sub-score evaluates whether the tone and expression of the message are likely to be accepted by the driver, using an ensemble of rule-based linguistic feature analysis and LLM Judge evaluation.

\subsubsection*{Rule-based score $S_\text{accept}^\text{rule}$} 
Linguistic features are detected through POS and syntactic analysis using spaCy, and penalties or bonuses are added to a base score of 0.5: imperative verbs (sentence-initial VB tags, e.g., stop/pull over/don't) incur $-0.3$ per occurrence, strong directive expressions (e.g., immediately/must/now) incur $-0.2$ per occurrence, and suggestive/interrogative expressions (e.g., would you/consider/might want) receive $+0.2$ per occurrence:
\begin{equation}\label{eq:s_accept_rule}
S_\text{accept}^\text{rule} = \text{clip}\left(0.5 + \sum \text{penalties/bonuses},\ 0,\ 1\right)
\end{equation}

\subsubsection*{LLM Judge score $S_\text{accept}^\text{judge}$} 
The GPT-5 Judge rates three aspects, ``non-imperative,'' ``non-intimidating,'' and ``suggestive style'', on a 1--5 scale, and the mean is normalized to $[0,1]$:
\begin{equation}\label{eq:s_accept_judge}
S_\text{accept}^\text{judge} = \frac{\text{mean}(\text{accept\_scores}) - 1}{4}
\end{equation}

\subsubsection*{Composite score}
The two components are combined with an ensemble weight $\beta$:
\begin{equation}\label{eq:s_accept}
S_\text{accept} = \beta \cdot S_\text{accept}^\text{rule} + (1 - \beta) \cdot S_\text{accept}^\text{judge}
\end{equation}
$\beta = 0.4$ is adopted. The rule-based component is deterministically reproducible, while the Judge component captures contextual nuances such as sarcastic polite expressions and culturally inappropriate phrasing.

\subsubsection{DSAIS Composite Score}

The five sub-scores are integrated by equal-weight arithmetic mean:
\begin{equation}\label{eq:dsais}
\text{DSAIS} = \frac{1}{5}\left(S_\text{risk} + S_\text{context} + S_\text{clarity} + S_\text{load} + S_\text{accept}\right)
\end{equation}
$\text{DSAIS} \in [0, 1]$, with higher values indicating intervention messages that are superior across safety, clarity, and acceptability. Equal weighting is adopted for two reasons: i)~there is currently insufficient empirical basis for establishing a priori priorities among the five quality dimensions, and ii)~unequal weighting introduces additional hyperparameters with a risk of overfitting to a specific dataset. Algorithm~\ref{alg:pipeline} summarizes the intervention generation pipeline, and Algorithm~\ref{alg:dsais} details the DSAIS evaluation procedure.

\begin{algorithm}[t]
\caption{Intervention Generation Pipeline}
\label{alg:pipeline}
\begin{algorithmic}[1]
\Require Recognition outputs $\{(y_k, c_k)\}_{k \in \mathcal{K}}$, history buffer $H$, session profile $P$
\Ensure Intervention message $m$, integrated risk $R$
\State Compute $r_k \gets d_k(y_k) \cdot c_k$ for each $k \in \mathcal{K}$
\State Compute $R \gets \sum_{k} w_k \cdot r_k$
\State Classify risk level: $\ell \gets \text{Level}(R)$
\State Update history $H \gets H \cup \{X_t\}$; compute $n_\text{persist}$
\State Update session profile $P$
\State Construct structured prompt from $R$, $\ell$, $H$, $P$
\State Generate message: $m, \hat{t}, \hat{\ell} \gets \text{LLM}(\text{prompt})$
\State \Return $m$, $R$
\end{algorithmic}
\end{algorithm}

\begin{algorithm}[t]
\caption{DSAIS Evaluation Procedure}
\label{alg:dsais}
\begin{algorithmic}[1]
\Require Message $m$, integrated risk $R$, task labels $\{y_k\}$, risk level $\ell$, ensemble weight $\beta$
\Ensure DSAIS score, sub-score vector
\State $t_\text{judge} \gets \text{Judge}_\text{tone}(m)$ \Comment{Tone classification: gentle/warning/urgent $\to$ $\{0, 0.5, 1.0\}$}
\State $S_\text{risk} \gets \max(0,\, \min(1,\, 1 - |t_\text{judge} - R|))$
\State $S_\text{context} \gets (\text{Judge}_\text{context}(m, \{y_k\}, \ell) - 1) / 4$ \Comment{1--5 scale $\to$ $[0, 1]$}
\State $W \gets \text{word\_count}(m)$
\State $S_\text{clarity} \gets \max(0,\, 1 - \max(0,\, W - W_\text{opt}) / (W_\text{max} - W_\text{opt}))$
\State $I \gets \text{info\_elements}(m)$ \Comment{Numerals + named entities + deictic expressions}
\State $S_\text{load} \gets \max(0,\, 1 - I / I_\text{max})$
\State $S_\text{accept}^\text{rule} \gets \text{clip}(0.5 + \sum \text{penalties/bonuses},\, 0,\, 1)$
\State $S_\text{accept}^\text{judge} \gets (\text{mean}(\text{Judge}_\text{accept}(m)) - 1) / 4$
\State $S_\text{accept} \gets \beta \cdot S_\text{accept}^\text{rule} + (1-\beta) \cdot S_\text{accept}^\text{judge}$
\State $\text{DSAIS} \gets \frac{1}{5}(S_\text{risk} + S_\text{context} + S_\text{clarity} + S_\text{load} + S_\text{accept})$
\State \Return DSAIS, $(S_\text{risk}, S_\text{context}, S_\text{clarity}, S_\text{load}, S_\text{accept})$
\end{algorithmic}
\end{algorithm}
\section{Experiments and Analysis}\label{sec:experiments}

\subsection{Experimental Design}\label{sec:experimental_design}

\subsubsection{Dataset}\label{sec:dataset}

Experiments were conducted using the AIDE dataset~\citep{3}. AIDE contains 3,062 driving video clips, each annotated with four task labels (TCR, VCR, DER, DBR). Each clip comprises footage from four camera views (front, in-cabin, left, and right). As preprocessing, 16 frames were uniformly sampled from each clip, view images were resized to $224 \times 224$, face crops to $64 \times 64$, and body crops to $112 \times 112$, followed by normalization with ImageNet statistics. Following the official split, the test set contains 609 samples. A key characteristic of AIDE is that each clip represents an independent short-duration driving scene with a single fixed label set per clip. That is, changes in driver state within a clip are not recorded, and each sample functions as a snapshot of a particular moment of driver state.

CauPsi (four-task average accuracy 82.7\%) was used to obtain predicted labels and confidence scores for all test samples. Inference was performed on an NVIDIA H100 NVL GPU at fp16 precision with a batch size of 4. Stratified sampling based on the integrated risk score $R$ was applied to the test set, yielding 213 evaluation samples as shown in Table~\ref{tab:stratification}. The limitation of only 13 HIGH-risk samples constrains statistical power; however, this reflects the natural characteristic of the AIDE dataset in which hazardous driving situations are rare.

\begin{table}[t]
\centering
\caption{Stratified sampling results by risk level.}
\label{tab:stratification}
\begin{tabular}{lcc}
\toprule
Risk Level & Full Test Set & Selected \\
\midrule
LOW ($R < 0.3$) & 383 & 100 \\
MEDIUM ($0.3 \leq R < 0.6$) & 213 & 100 \\
HIGH ($R \geq 0.6$) & 13 & 13 \\
\midrule
Total & 609 & 213 \\
\bottomrule
\end{tabular}
\end{table}

Constructing the state history $H$ and session profile $P$ defined in Section~\ref{sec:history} requires temporally continuous driving records. However, each AIDE sample is an independent video clip with no temporal continuity between samples. To address this constraint, the 213 selected samples were arranged in their order of appearance in the test set CSV to form a pseudo-session for state history construction. The rationale for this design is twofold. First, since each AIDE clip has a single fixed label set with no state changes within a clip, sequentially arranging independent clips generates a sequence of different label combinations over time, approximately simulating continuous driving in real-world operation. Second, the framework input consists solely of arrays of predicted labels and confidence scores, without using individual driver identity or video visual content; therefore, the fact that drivers differ between samples does not affect framework operation. Session profiles were reconstructed on the 213 selected samples, with profiles built from ground-truth labels for the GT condition and from predicted labels for the pred condition to ensure consistency across conditions.

\subsubsection{Generation Models and Baselines}\label{sec:models}

Table~\ref{tab:models} lists the models used for message generation. Four open-source LLMs and one API-based model were selected to ensure diversity in developer, architecture, and training data. All local models were run on an NVIDIA H100 NVL GPU (95 GB) at fp16 precision without quantization. By not applying quantization, the confounding factor of quantization-induced accuracy degradation is eliminated, enabling accurate comparison of each model's intrinsic performance.

\begin{table}[t]
\centering
\caption{List of message generation models.}
\label{tab:models}
\small
\makebox[\textwidth][c]{%
\begin{tabular}{lccl}
\toprule
Model & Parameters & Type & Selection Rationale \\
\midrule
Qwen2.5-7B-Instruct & 7B & Local & High multilingual performance; deployable in-vehicle \\
Qwen2.5-14B-Instruct & 14B & Local & Effect of model size \\
Llama-3.1-8B-Instruct & 8B & Local & Different architecture \\
Gemma-2-9B-it & 9B & Local & Different architecture \\
GPT-5-mini & --- & API & API performance upper bound \\
\bottomrule
\end{tabular}}
\end{table}

Since the problem setting of LLM intervention message generation from four-task integrated context is itself novel, no directly comparable existing methods exist. Therefore, two baselines are established to validate the design decisions of the proposed framework: i)~a rule-based system that randomly selects from three predefined templates per risk level, and ii)~a single-task (DER only) baseline that uses only the DER output and computes a DER-specific risk score. The rule-based baseline represents the performance floor without LLM use, while the single-task baseline corresponds to the single-task input condition addressed by prior research. The single-task baseline does not use the four-task integrated $R$ but injects only the DER-specific risk score and risk level into the prompt, preventing multi-task information leakage.

\subsubsection{Experimental Conditions}\label{sec:experimental_conditions}

To evaluate robustness against upstream recognition errors, seven conditions were defined as shown in Table~\ref{tab:conditions}. Based on the two input conditions defined in the problem formulation, planned misclassification simulations from $X_\text{gt}$ are added to systematically analyze the impact of recognition error type and count on intervention quality.

\begin{table}[t]
\centering
\caption{List of experimental conditions. In misclassification conditions, a label other than the ground truth is randomly selected, and confidence is sampled from $U(0.4, 0.8)$.}
\label{tab:conditions}
\makebox[\textwidth][c]{%
\begin{tabular}{ll}
\toprule
Condition & Description \\
\midrule
GT & Ground-truth labels (confidence 1.0) \\
1err-TCR & TCR misclassified (other 3 tasks correct) \\
1err-VCR & VCR misclassified (other 3 tasks correct) \\
1err-DER & DER misclassified (other 3 tasks correct) \\
1err-DBR & DBR misclassified (other 3 tasks correct) \\
2err & 2 tasks randomly and simultaneously misclassified \\
pred & Actual predictions and confidence scores from CauPsi \\
\bottomrule
\end{tabular}}
\end{table}

By separating the single-task misclassification conditions (1err) by task, it is possible to quantitatively analyze which task's misclassification most severely affects intervention message quality. The pred condition uses actual recognition model outputs rather than simulations, thereby directly evaluating overall performance under real-world operating conditions.

All local model inference was performed using the Hugging Face Transformers library with AutoModelForCausalLM at fp16 precision. Seed 42 was fixed across all experiments to ensure reproducibility. As Gemma-2-9B-it does not support the system role, a fallback was implemented that integrates the system prompt at the beginning of the user message.

For local models, temperature 0.3 and max\_tokens 150 were used as common settings. For GPT-5-mini, temperature was run at default settings. max\_completion\_tokens was set to 1000 because GPT-5-series models internally consume reasoning tokens, and a setting of 150 causes all tokens to be spent on reasoning, resulting in empty outputs. The GPT-5 Judge was similarly run at default settings.

\subsubsection{Evaluation Metrics}\label{sec:metrics}

In addition to DSAIS and its five sub-scores, the following auxiliary metrics are used. BERTScore F1 measures the semantic similarity between GT condition and each condition's messages. Because BERTScore's discriminative power may decrease for short texts of 15 words or fewer, the Structural Match Rate (SMR) is used as a complementary metric.

SMR is the proportion of samples in which both the tone and risk level fields exactly match between the GT condition and each condition:
\begin{equation}\label{eq:smr}
\text{SMR}_k = \frac{1}{N} \sum_{i=1}^{N} \mathbbm{1}\left[\text{tone}_k^{(i)} = \text{tone}_\text{GT}^{(i)} \wedge \text{risk\_level}_k^{(i)} = \text{risk\_level}_\text{GT}^{(i)}\right]
\end{equation}

\subsubsection{DSAIS Validation Design}\label{sec:dsais_validation_design}

To validate the reliability of DSAIS from multiple perspectives, two additional experiments are conducted. For multi-judge agreement analysis, to verify that DSAIS scores are not artifacts of a single Judge model, GT condition messages from Qwen2.5-7B (213 samples) are evaluated by three architecturally distinct Judges shown in Table~\ref{tab:judges}. Kendall's $\tau$ (rank correlation) and Intraclass Correlation Coefficient (ICC) are computed for both $S_\text{context}$ and $S_\text{accept}$. ICC is interpreted following Cicchetti's guidelines~\citep{39}: ICC $< 0.40$ as poor, 0.40--0.59 as fair, 0.60--0.74 as good, and 0.75 or above as excellent.

\begin{table}[t]
\centering
\caption{Models used for multi-judge evaluation.}
\label{tab:judges}
\begin{tabular}{llll}
\toprule
Judge & Model & Type \\
\midrule
Primary Judge & GPT-5 & API \\
Secondary Judge 1 & Claude Sonnet 4 & API \\
Secondary Judge 2 & Qwen2.5-32B-Instruct & Local \\
\bottomrule
\end{tabular}
\end{table}

For discriminative power validation through control conditions, five control conditions with intentionally manipulated quality are constructed to verify that DSAIS can appropriately discriminate quality differences: i) Gold (high-quality samples filtered by $S_\text{risk} \geq 0.8$, $S_\text{clarity} \geq 0.8$, and $S_\text{load} \geq 0.8$), ii) Mismatch-tone (gentle tone forced for HIGH-risk), iii) Overlong (substantially exceeding the 15-word limit), iv) Imperative (imperative messages), and v) Random (contextually irrelevant sentences). Each condition comprises 50 samples (Mismatch-tone is limited to 13 samples due to the constraint on HIGH-risk sample count), and Cohen's $d$ against Gold is computed. The target is to achieve $d \geq 0.8$ (large effect size) across all conditions.


\subsection{DSAIS Validation}\label{sec:dsais_validation}

\subsubsection{Multi-Judge Agreement}

Table~\ref{tab:multi_judge} presents the results of the multi-judge agreement analysis. ICC(C,k) = 0.798 for $S_\text{context}$ is classified as ``good agreement'' and ICC(C,k) = 0.840 for $S_\text{accept}$ as ``excellent agreement.'' The high agreement across three architecturally distinct models supports that DSAIS evaluation is a robust metric not dependent on a specific LLM.

\begin{table}[t]
\centering
\caption{Results of multi-judge agreement analysis.}
\label{tab:multi_judge}
\small
\begin{tabular}{lcccc}
\toprule
Sub-score & \multicolumn{2}{c}{Kendall's $\tau$} & \multicolumn{2}{c}{ICC(C,k)} \\
\cmidrule(lr){2-3} \cmidrule(lr){4-5}
 & GPT-5 vs Claude & GPT-5 vs Qwen-32B & ICC & 95\% CI \\
\midrule
$S_\text{context}$ & 0.458 & 0.454 & 0.798 & [0.75, 0.84] \\
$S_\text{accept}$ & 0.577 & 0.548 & 0.840 & [0.80, 0.87] \\
\bottomrule
\end{tabular}
\end{table}

The higher agreement for $S_\text{accept}$ than for $S_\text{context}$ is attributable to the fact that judging whether a message is suggestive or imperative is a relatively surface-level linguistic evaluation, whereas assessing contextual relevance requires reasoning about the relationship between driver state and message content, constituting a more subjective judgment.

Note that ICC(1,1) (absolute agreement of a single Judge) is 0.333 for $S_\text{accept}$, which is somewhat low; this reflects differences in scoring scale (leniency/strictness) across Judges. ICC(C,k) (consistency-based, averaged across 3 Judges) is 0.840, confirming that ranking consistency is maintained.

\subsubsection{Sanity Check with Control Conditions}

Five control conditions with intentionally manipulated quality were constructed to validate the discriminative power of DSAIS. Table~\ref{tab:control} presents the results. Cohen's $d > 1.5$ was achieved across all conditions, substantially exceeding the pre-specified target of $d \geq 0.8$. The largest effect size for Mismatch-tone ($d = 6.90$; $n = 13$) is attributable to the forced gentle tone for HIGH-risk directly reducing $S_\text{risk}$, though this estimate should be interpreted with caution given the small sample size, which may inflate Cohen's $d$. Overlong and Imperative are primarily driven by decreases in $S_\text{clarity}$ and $S_\text{accept}$, respectively, confirming that each sub-score captures its intended quality dimension. The relatively smaller effect size for Random ($d = 1.52$) is because irrelevant sentences do not affect $S_\text{clarity}$ or $S_\text{load}$, with the decrease concentrated in $S_\text{context}$.

\begin{table}[t]
\centering
\caption{Control condition results.}
\label{tab:control}
\begin{tabular}{lccc}
\toprule
Condition & DSAIS (mean) & Cohen's $d$ & $n$ \\
\midrule
Gold & 0.821 & --- & 50 \\
Mismatch-tone & 0.569 & 6.90 & 13 \\
Overlong & 0.567 & 4.96 & 50 \\
Imperative & 0.541 & 3.78 & 50 \\
Random & 0.677 & 1.52 & 50 \\
\bottomrule
\end{tabular}
\end{table}

\subsection{Framework Effectiveness and Cross-Model Comparison}\label{sec:rq1}

Table~\ref{tab:comparison} presents the results for all methods under the GT condition. The rule-based baseline and single-task (DER) baseline are included to contextualize the proposed framework's performance. The rule-based baseline was established to verify the necessity of LLM-based intervention. Although it recorded the highest overall DSAIS, this is attributable to structurally perfect scores on $S_\text{clarity}$ and $S_\text{load}$ from its short, fixed templates, and does not reflect capability as an intervention system. Its $S_\text{context}$ remained low, confirming that LLMs are indispensable for generating context-adaptive intervention messages.

\begin{table}[t]
\centering
\caption{Comparison of all methods (GT condition). $p$-values are from pairwise comparison with Qwen2.5-7B using the Wilcoxon signed-rank test. No multiple comparison correction was applied; however, all significant results survive Bonferroni correction (adjusted $\alpha = 0.05/4 = 0.0125$).}
\label{tab:comparison}
\small
\makebox[\textwidth][c]{%
\begin{tabular}{lcccccccc}
\toprule
Method & DSAIS & $\sigma$ & $S_\text{risk}$ & $S_\text{ctx}$ & $S_\text{clr}$ & $S_\text{load}$ & $S_\text{acc}$ & $p$ \\
\midrule
Rule-based & 0.795 & 0.088 & 0.708 & 0.669 & 1.000 & 1.000 & 0.598 & --- \\
Single-task (DER) & 0.774 & 0.081 & 0.685 & 0.711 & 0.994 & 1.000 & 0.481 & --- \\
\midrule
Qwen2.5-7B & 0.774 & 0.062 & 0.686 & 0.730 & 0.949 & 0.998 & 0.507 & --- \\
Qwen2.5-14B & 0.763 & 0.088 & 0.685 & 0.590 & 0.822 & 0.994 & 0.725 & 0.128 \\
Llama-3.1-8B & 0.685 & 0.083 & 0.684 & 0.545 & 0.700 & 0.995 & 0.501 & $<$0.001$^{***}$ \\
Gemma-2-9B & 0.790 & 0.072 & 0.680 & 0.673 & 0.977 & 1.000 & 0.623 & 0.004$^{**}$ \\
GPT-5-mini & 0.653 & 0.073 & 0.696 & 0.516 & 0.323 & 0.961 & 0.768 & $<$0.001$^{***}$ \\
\bottomrule
\end{tabular}}
\end{table}

\begin{figure}[t]
\centering
\includegraphics[width=0.8\columnwidth]{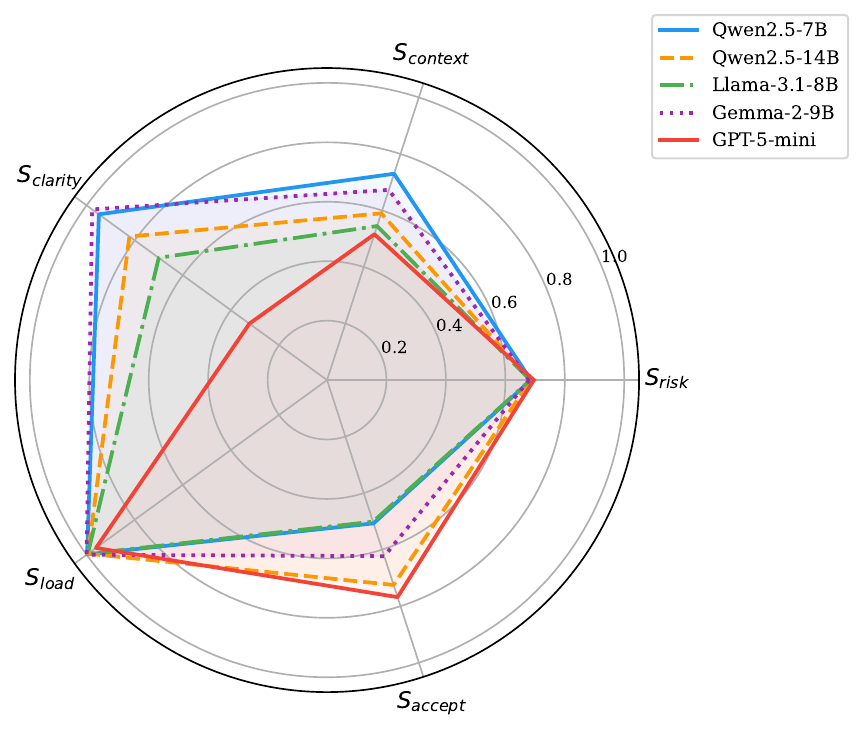}
\caption{Sub-score profiles of five generation models (GT condition).}
\label{fig:radar}
\end{figure}

Figure~\ref{fig:radar} visualizes the sub-score profiles of all five models as a radar chart, clearly illustrating the distinct quality characteristics of each model.

Among LLM-based methods, Qwen2.5-7B achieved the highest $S_\text{context}$, representing a 2.7\% improvement over the single-task baseline. $S_\text{risk}$ and $S_\text{load}$ also maintained high values, confirming that four-task integration enables accurate situational understanding while generating messages with risk-aligned tone and low cognitive load. Furthermore, the standard deviation was the smallest among all conditions ($\sigma = 0.062$), demonstrating robust and stable intervention quality across diverse inputs.

Gemma-2-9B recorded the highest overall DSAIS, exhibiting a balanced profile across all sub-scores. Notably, Qwen2.5-14B exhibited lower $S_\text{context}$ than the 7B variant. While increasing model size contributes to improved $S_\text{accept}$, it does not directly translate to improved contextual relevance. The Wilcoxon signed-rank test revealed no significant difference between the two Qwen variants ($p = 0.128$), while differences with all other models were statistically significant.

GPT-5-mini recorded the lowest DSAIS, primarily due to low $S_\text{clarity}$. Its messages average 12--14 words compared to 5--8 words for local models, reflecting a tendency toward more detailed and explanatory outputs attributable to its reasoning token architecture. Conversely, $S_\text{accept}$ was the highest among all models, indicating superior capability in generating acceptable expressions. These results demonstrate that compact local models (7B--9B) can achieve intervention quality equal to or superior to large-scale API-based models, supporting the feasibility of on-device deployment for in-vehicle systems.

\subsection{Robustness to Misclassification}\label{sec:rq3}

Table~\ref{tab:bertscore} presents the full BERTScore and SMR results. A consistent pattern was observed across all models: DER misclassification caused the largest change in message content. For GPT-5-mini, SMR dropped to 0.484 under DER misclassification, meaning that tone or risk\_level classification changed in more than half of the generated messages. This represents the largest single-condition degradation observed across all models and conditions.

\begin{table}[!t]
\centering
\caption{BERTScore F1 and SMR results across models and conditions.}
\label{tab:bertscore}
\small
\begin{tabular}{llcc}
\toprule
Model & Condition & BERTScore F1 & SMR \\
\midrule
\multirow{6}{*}{Qwen2.5-7B} & 1err-TCR & 0.925 & 0.864 \\
 & 1err-VCR & 0.930 & 0.829 \\
 & 1err-DER & 0.899 & 0.751 \\
 & 1err-DBR & 0.907 & 0.765 \\
 & 2err & 0.897 & 0.704 \\
 & pred & 0.922 & 0.751 \\
\midrule
\multirow{6}{*}{Qwen2.5-14B} & 1err-TCR & 0.969 & 0.845 \\
 & 1err-VCR & 0.962 & 0.791 \\
 & 1err-DER & 0.939 & 0.535 \\
 & 1err-DBR & 0.937 & 0.819 \\
 & 2err & 0.939 & 0.634 \\
 & pred & 0.950 & 0.711 \\
\midrule
\multirow{6}{*}{Llama-3.1-8B} & 1err-TCR & 0.941 & 0.906 \\
 & 1err-VCR & 0.939 & 0.869 \\
 & 1err-DER & 0.917 & 0.751 \\
 & 1err-DBR & 0.926 & 0.890 \\
 & 2err & 0.926 & 0.801 \\
 & pred & 0.930 & 0.831 \\
\midrule
\multirow{6}{*}{Gemma-2-9B} & 1err-TCR & 0.922 & 0.838 \\
 & 1err-VCR & 0.938 & 0.758 \\
 & 1err-DER & 0.909 & 0.601 \\
 & 1err-DBR & 0.913 & 0.789 \\
 & 2err & 0.907 & 0.660 \\
 & pred & 0.926 & 0.749 \\
\midrule
\multirow{6}{*}{GPT-5-mini} & 1err-TCR & 0.899 & 0.878 \\
 & 1err-VCR & 0.901 & 0.784 \\
 & 1err-DER & 0.895 & 0.484 \\
 & 1err-DBR & 0.897 & 0.836 \\
 & 2err & 0.896 & 0.643 \\
 & pred & 0.900 & 0.718 \\
\bottomrule
\end{tabular}
\end{table}

\begin{figure}[t]
\centering
\includegraphics[width=\columnwidth]{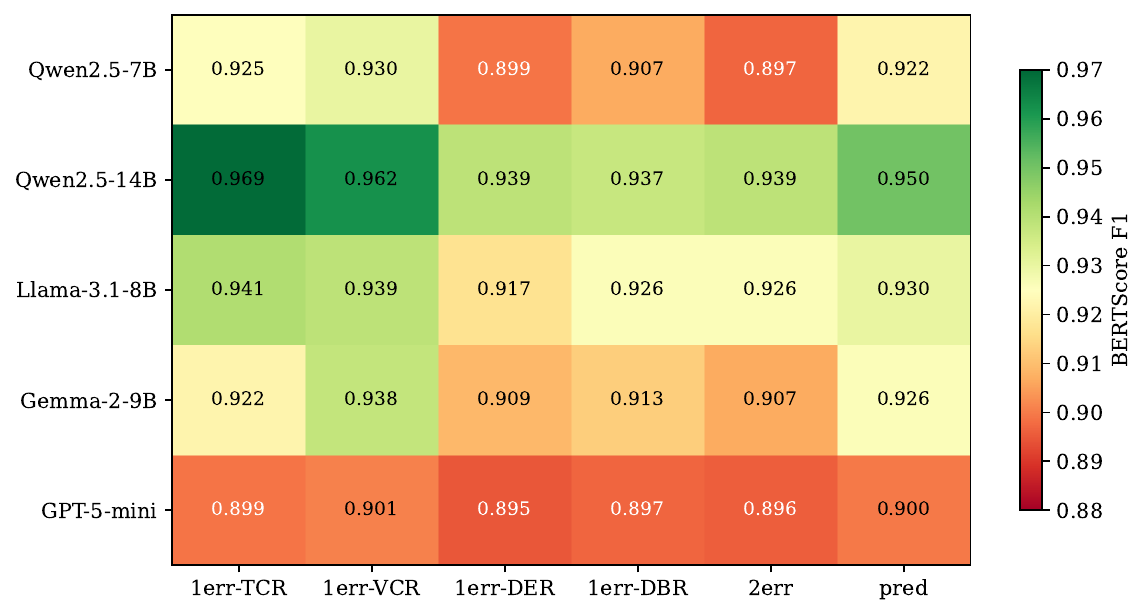}
\caption{BERTScore F1 heatmap across models and misclassification conditions. Darker colors indicate lower similarity to GT condition messages.}
\label{fig:heatmap}
\end{figure}

Figure~\ref{fig:heatmap} visualizes the BERTScore F1 across all models and conditions as a heatmap, highlighting the consistent DER vulnerability pattern.

This result is intuitively interpretable. TCR and VCR are descriptive information about the external environment; even when misrecognized, the fundamental direction of intervention does not change. In contrast, DER misrecognition alters the core intent of the message. For example, misclassifying Weariness as Anger transforms a rest suggestion into an emotional regulation suggestion, which could be counterproductive for a fatigued driver.

Focusing on the pred condition, BERTScore F1 exceeded 0.900 for all models, indicating that messages close to the GT condition are generated even when using actual recognition model outputs. SMR was 0.751 for Qwen2.5-7B, 0.711 for Qwen2.5-14B, and 0.831 for Llama-3.1-8B, comparable to the 1err-DER condition. This suggests that the framework maintains practical intervention quality under real-world operating conditions with CauPsi's average accuracy of 82.7\%.

Qwen2.5-14B recorded the highest BERTScore F1 across all conditions (minimum 0.937), yet its SMR under DER misclassification was the lowest among all models (0.535). This divergence indicates that the 14B model generates semantically similar surface text regardless of input perturbation but fails to adjust tone and risk metadata accordingly, consistent with the lower $S_\text{context}$ observed in Table~\ref{tab:comparison}. Llama-3.1-8B exhibited the highest robustness on the SMR dimension, maintaining BERTScore F1 of 0.917 even under DER misclassification with substantially higher SMR (0.751) than the 14B variant. This suggests that Llama's generation strategy is less sensitive to individual input perturbations. The two-task simultaneous misclassification condition (2err) yielded the lowest scores across all models, confirming the cumulative nature of upstream error propagation.

\subsection{Tone Self-Assessment Analysis}\label{sec:tone_self}

Table~\ref{tab:tone} presents the agreement rate between each model's self-reported tone and the GPT-5 Judge's tone classification. Llama-3.1-8B achieved a 97.2\% agreement rate, substantially outperforming all other models. This result indicates that Llama-3.1-8B is highly accurate in self-assessing output tone, suggesting that tone metadata can be trusted without an external Judge during in-vehicle operation. In contrast, Qwen2.5-14B and GPT-5-mini remained at approximately 50\% agreement, confirming that the reliability of self-reported tone varies substantially across models. The JSON output failure rate was 0.0\% (0 out of 213) for all models, confirming the reliability of structured output generation.

\begin{table}[t]
\centering
\caption{Tone self-assessment agreement rate and Cohen's $\kappa$ (GT condition).}
\label{tab:tone}
\begin{tabular}{lcc}
\toprule
Model & Agreement & $\kappa$ \\
\midrule
Qwen2.5-7B & 0.704 & $-$0.102 \\
Qwen2.5-14B & 0.484 & 0.000 \\
Llama-3.1-8B & \textbf{0.972} & \textbf{0.246} \\
Gemma-2-9B & 0.718 & $-$0.047 \\
GPT-5-mini & 0.507 & 0.071 \\
\bottomrule
\end{tabular}
\end{table}

The addition of Cohen's $\kappa$ reveals a critical distinction between raw agreement and chance-corrected agreement. While Qwen2.5-7B achieves 70.4\% raw agreement, its $\kappa = -0.102$ indicates performance below chance, attributable to the dominance of a single tone category (gentle) in both self-reports and Judge classifications. Only Llama-3.1-8B achieves a positive $\kappa$ (0.246, fair agreement), confirming that its high raw agreement reflects genuine tone awareness rather than distributional coincidence. These results suggest that for practical deployment, tone self-assessment should not be trusted without external validation, except for Llama-3.1-8B.

\subsection{Framework Ablation Study}\label{sec:ablation}

To quantify the contribution of each framework component, ablation experiments were conducted by systematically removing individual modules from the full framework. All ablation experiments used Qwen2.5-7B under the GT condition to isolate framework effects from model and recognition variability.

\begin{figure}[t]
\centering
\includegraphics[width=\columnwidth]{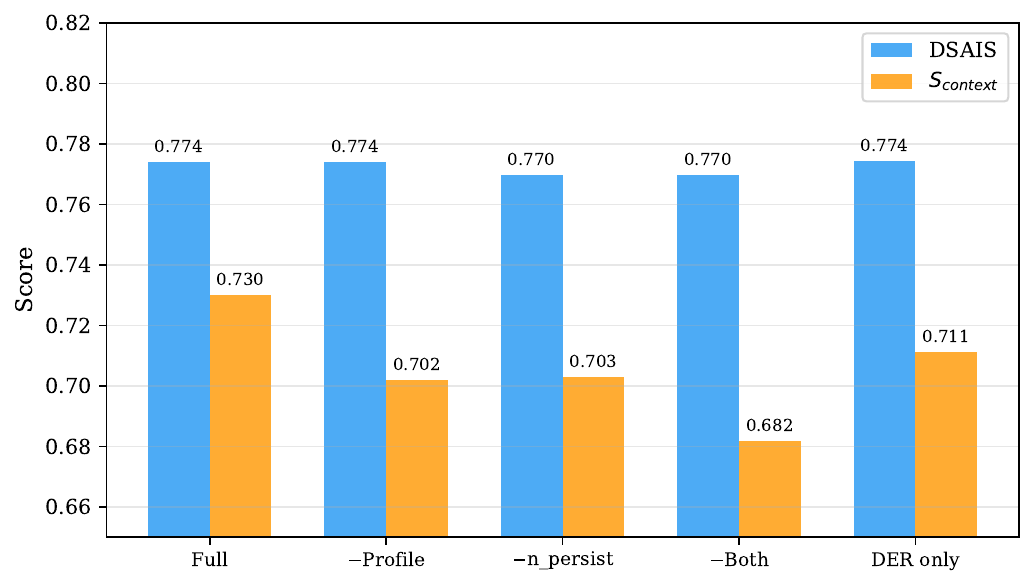}
\caption{Framework ablation: overall DSAIS vs.\ $S_\text{context}$ across configurations (Qwen2.5-7B, GT condition).}
\label{fig:ablation}
\end{figure}

Figure~\ref{fig:ablation} contrasts overall DSAIS with $S_\text{context}$ across five configurations. While overall DSAIS remains stable across all configurations (0.770--0.774), $S_\text{context}$ reveals clear differentiation: the full framework achieves the highest $S_\text{context}$ (0.730), followed by the single-task baseline (0.711), then configurations with removed components ($-n_\text{persist}$: 0.703; $-$Profile: 0.702; $-$Both: 0.682). This pattern demonstrates that each framework component contributes to contextual relevance, while the overall DSAIS stability reflects a compensating mechanism in which richer contextual input produces slightly longer messages, and the equal-weight aggregation absorbs the trade-off. Sub-score decomposition is therefore essential for evaluating individual component contributions.

Notably, removing Session Profile decreases $S_\text{context}$ by 3.8\%, and removing both Session Profile and $n_\text{persist}$ decreases it by 6.6\%, while overall DSAIS remains virtually unchanged. The moderate effect of Session Profile under these conditions is partly attributable to the AIDE dataset structure: each clip is an independent snapshot with no temporal continuity, limiting the discriminative value of session-level personalization features. With continuous driving data where genuine state transitions occur, session-level components are expected to provide greater benefit.

\subsection{Sensitivity Analysis}\label{sec:sensitivity}

To evaluate the robustness of the proposed design, sensitivity analyses were conducted on task weight allocation and the acceptability ensemble weight $\beta$. Switching from the proposed task weights to uniform weights ($w_k = 0.25$) produces only a 0.7\% DSAIS difference (0.774 vs.\ 0.777), confirming that overall framework performance is robust to reasonable weight variations.

Table~\ref{tab:beta} presents the effect of the ensemble weight $\beta$ in $S_\text{accept}$ (Eq.~\ref{eq:s_accept}) across the range $[0, 1]$.

\begin{table}[t]
\centering
\caption{Effect of acceptability ensemble weight $\beta$ on DSAIS (Qwen2.5-7B, GT condition).}
\label{tab:beta}
\begin{tabular}{ccc}
\toprule
$\beta$ & DSAIS (mean) & $\sigma$ \\
\midrule
0.0 (Judge only) & 0.799 & 0.066 \\
0.2 & 0.787 & 0.064 \\
0.4 (Adopted) & 0.774 & 0.062 \\
0.6 & 0.762 & 0.062 \\
0.8 & 0.749 & 0.062 \\
1.0 (Rule only) & 0.737 & 0.064 \\
\bottomrule
\end{tabular}
\end{table}

DSAIS decreases monotonically as $\beta$ increases from 0 (Judge only) to 1 (rule only), with the Judge-only configuration yielding the highest DSAIS. The adopted value of $\beta = 0.4$ balances two considerations: i)~reproducibility, since the rule-based component is deterministic while the Judge component depends on API availability and version, and ii)~contextual sensitivity, since the Judge captures nuances such as sarcastic politeness and culturally inappropriate phrasing that rule-based detection cannot identify. The standard deviation is minimized at $\beta = 0.4$--$0.6$, indicating that the ensemble stabilizes evaluation. The monotonic trend confirms that the LLM Judge consistently rates messages more favorably than the rule-based detector, which tends to penalize any imperative-like construction regardless of contextual appropriateness.

\section{Discussion}\label{sec:discussion}

\subsection{Insights Enabled by DSAIS}

The multi-dimensional evaluation enabled by DSAIS revealed several insights that conventional NLG metrics cannot capture. The $S_\text{context}$ sub-score quantified the fundamental limitation of rule-based systems: although the rule-based baseline recorded the highest overall DSAIS (0.795) owing to structurally perfect $S_\text{clarity}$ and $S_\text{load}$, its $S_\text{context}$ (0.669) was surpassed by LLMs that leverage multi-task integration (Qwen2.5-7B: 0.730, Gemma-2-9B: 0.673). DSAIS thus distinguishes messages that are concise but lack situational understanding from those that accurately reflect the driving context, providing quantitative evidence that multi-task LLM-based generation can achieve superior contextual relevance.

Sub-score profile analysis also enabled multi-faceted model characterization. GPT-5-mini achieved the highest $S_\text{accept}$ (0.768) yet the lowest $S_\text{clarity}$ (0.323), revealing a verbose tendency attributable to its reasoning token architecture. Qwen2.5-14B outperformed the 7B variant on $S_\text{accept}$ (0.725 vs.\ 0.507) but underperformed on $S_\text{context}$ (0.590 vs.\ 0.730), plausibly because the larger model generates more elaborate and hedged language that the Judge perceives as less situation-specific. These trade-offs are invisible from the aggregate DSAIS and demonstrate the value of sub-score profiles for model selection.

Misclassification experiments revealed that DER accuracy is the most critical upstream factor: for GPT-5-mini, SMR dropped to 0.484 under DER misclassification, far below TCR (0.878) or DBR (0.836). Emotion misrecognition alters the core intent of the message, whereas traffic or vehicle context misrecognition affects descriptive information without changing the intervention direction. A methodological caveat is that DER carries the highest task weight ($w_\text{DER} = 0.35$), which mechanically amplifies its influence; however, the sensitivity analysis shows that switching to uniform weights produces only a 0.7\% DSAIS difference, suggesting that the DER effect is robust.

The ablation study demonstrated that each framework component contributes to contextual relevance, with $S_\text{context}$ decreasing from 0.730 to 0.702 without Session Profile and to 0.682 without both Session Profile and $n_\text{persist}$, while overall DSAIS remains stable. This diagnostic capability illustrates a key strength of DSAIS: sub-score decomposition reveals component-level contributions that aggregate scoring masks.

\subsection{Design Guidelines for Practical Systems}

Existing LLM-based driver assistance systems such as CAWA~\citep{45} and Talk2Drive~\citep{46} have explored context-adaptive warnings and natural language commands for vehicle control, respectively, but neither generates free-text intervention messages from multi-task recognition outputs or provides a domain-specific metric for evaluating message quality. The systematic experiments in this study, which uniquely combine four-task integration with multi-dimensional evaluation, yield the following design guidelines that extend beyond what prior work has addressed.

Upstream recognition systems should allocate computational resources disproportionately to emotion recognition. DER misclassification caused the largest degradation across all models (SMR as low as 0.484), because emotion misrecognition alters the core intent of the intervention message, unlike traffic or vehicle context errors that affect only descriptive information. This finding provides a concrete resource allocation criterion that single-modality systems such as CAWA cannot derive.

Compact local LLMs (7B--9B) achieve intervention quality equal to or superior to 14B and API-based models, making on-device deployment feasible without quality compromise. Llama-3.1-8B offers the highest robustness to upstream errors, while Qwen2.5-7B achieves the highest $S_\text{context}$. This model-level characterization through sub-score profiles enables informed selection based on deployment priorities, an evaluation capability absent from prior driver assistance frameworks.

Overall DSAIS alone is insufficient for model selection or quality assurance, as the ablation study demonstrated that component-level contributions are visible only through sub-score decomposition. For continuous in-vehicle monitoring, $S_\text{clarity}$ and $S_\text{load}$ can be computed in a lightweight rule-based manner without an external LLM Judge, providing a practical monitoring pathway.

Tone self-assessment should not be trusted without external validation. Only Llama-3.1-8B achieves positive chance-corrected agreement ($\kappa = 0.246$); for all other models, an external Judge or rule-based verification is required to ensure that the reported tone matches the actual message content.

\subsection{Limitations and Future Work}\label{sec:limitations}

The most significant constraint stems from the AIDE dataset, which consists of independent video clips rather than continuous driving sessions. While Session Profile removal still decreases $S_\text{context}$ by 3.8\%, the full potential of session-level personalization is expected to manifest more clearly with temporally continuous data. Additionally, only 13 HIGH-risk samples constrain statistical power for safety-critical scenarios. Validation on continuous driving datasets~\citep{41} and data collection focused on high-risk scenarios are essential future directions.

This study relies entirely on LLM Judges for $S_\text{context}$ and $S_\text{accept}$, and while multi-judge agreement is high (ICC up to 0.840), how actual drivers perceive messages and modify their behavior has not been verified. Usability studies and driving simulator experiments with human participants are the highest-priority future task. Multilingual evaluation is also an important direction toward practical deployment.

The ceiling effect of $S_\text{clarity}$ and $S_\text{load}$ (both near 1.0 for most models) limits the discriminative power of overall DSAIS. Potential remedies include harmonic mean aggregation and minimum-score-weighted aggregation. Sub-score weight optimization informed by human evaluation, adaptive scoring that accounts for risk level, and verification of quantization effects on intervention quality remain open questions.

\section{Conclusion}\label{sec:conclusion}

We proposed DSAIS, a domain-specific composite metric that evaluates five quality dimensions of driver intervention messages through a hybrid architecture combining rule-based computation with LLM Judge evaluation, together with an end-to-end framework integrating four-task recognition outputs via risk fusion and dynamic prompt construction. Experiments on the AIDE dataset with five models and seven conditions demonstrated inter-judge agreement of ICC 0.798--0.840, discriminative power of Cohen's $d > 1.5$, and a 9.1\% contextual relevance improvement over rule-based baselines. Ablation experiments confirmed that each framework component contributes to contextual relevance through sub-score decomposition that aggregate scoring masks. Misclassification experiments identified driver emotion recognition as the most critical upstream factor, and compact local LLMs (7B--9B) outperformed API-based models, providing practical guidelines for in-vehicle deployment.

Future work includes validation on continuous driving datasets, usability studies with human drivers, and sub-score weight optimization informed by human evaluation.

\section*{Declaration of competing interest}
The authors declare that they have no known competing financial interests or personal relationships that could have appeared to influence the work reported in this paper.

\section*{Acknowledgment}
The authors would like to express their sincere gratitude to Isuzu Motors Limited, Isuzu Advanced Engineering Center Limited, and NineSigma Holdings, Inc. for their valuable advice and insightful feedback throughout this study.

\section*{Data availability}
The dataset used in this study is publicly available at \url{https://github.com/ydk122024/AIDE}.

\section*{Declaration of Generative AI Use}
During the preparation of this manuscript, the authors used generative AI tools for language editing and organization. All content was reviewed and verified by the authors, who take full responsibility for the final manuscript.

\bibliographystyle{elsarticle-harv}
\bibliography{refs}

\clearpage

\appendix

\section{Generation Prompts}\label{app:prompts}

\subsection*{System Prompt (Multi-Task)}

The following system prompt is used for all LLM-based generation models in the proposed framework:

\begin{quote}
\small
You are an in-vehicle safety assistant AI.\\
Based on the driver state data provided, generate an intervention message following these constraints:\\
\hspace*{1em}- English, concise (15 words or fewer)\\
\hspace*{1em}- Avoid imperative tone; prefer suggestion/guidance style\\
\hspace*{1em}- Match urgency to the risk level (LOW/MEDIUM/HIGH)\\
\hspace*{1em}- Consider the driver session history for personalization\\
\hspace*{1em}- Output ONLY valid JSON: no preamble, no explanation\\
Output format:\\
\hspace*{1em}\{``message'': ``...'', ``tone'': ``gentle|warning|urgent'', ``risk\_level'': ``LOW|MEDIUM|HIGH''\}
\end{quote}

\subsection*{System Prompt (Single-Task Baseline)}

The single-task (DER only) baseline uses the following system prompt, which omits session history personalization:

\begin{quote}
\small
You are an in-vehicle safety assistant AI.\\
Based on the driver emotion data provided, generate an intervention message following these constraints:\\
\hspace*{1em}- English, concise (15 words or fewer)\\
\hspace*{1em}- Avoid imperative tone; prefer suggestion/guidance style\\
\hspace*{1em}- Match urgency to the risk level (LOW/MEDIUM/HIGH)\\
\hspace*{1em}- Output ONLY valid JSON: no preamble, no explanation\\
Output format:\\
\hspace*{1em}\{``message'': ``...'', ``tone'': ``gentle|warning|urgent'', ``risk\_level'': ``LOW|MEDIUM|HIGH''\}
\end{quote}

\subsection*{User Prompt Template (Multi-Task)}

The user prompt is dynamically populated per sample. Placeholder values are shown in braces:

\begin{quote}
\small
Driver state at sample \{sample\_id\}:\\
\hspace*{1em}TCR (Traffic): \ label=\{tcr\_label\}, \ conf=\{tcr\_conf\}, \ risk=\{tcr\_risk\}\\
\hspace*{1em}VCR (Vehicle): \ label=\{vcr\_label\}, \ conf=\{vcr\_conf\}, \ risk=\{vcr\_risk\}\\
\hspace*{1em}DER (Emotion): \ label=\{der\_label\}, \ conf=\{der\_conf\}, \ risk=\{der\_risk\}\\
\hspace*{1em}DBR (Behavior): label=\{dbr\_label\}, \ conf=\{dbr\_conf\}, \ risk=\{dbr\_risk\}\\
\hspace*{1em}Integrated risk score R = \{R\} -> \{risk\_level\}\\
\hspace*{1em}Persistent risk frames (last \{N\}): \{n\_persist\} / \{N\}\\[6pt]
{[}Session Profile{]}\\
\hspace*{1em}Drowsiness detected: \{drowsiness\_count\} times\\
\hspace*{1em}Dozing Off detected: \{dozing\_count\} times\\
\hspace*{1em}Phone use detected: \{phone\_count\} times\\
\hspace*{1em}Anger detected: \{anger\_count\} times\\[6pt]
Generate an intervention message in JSON format.
\end{quote}

\subsection*{User Prompt Template (Single-Task Baseline)}

\begin{quote}
\small
Driver emotion state:\\
\hspace*{1em}Emotion: label=\{der\_label\}, confidence=\{der\_conf\}\\
\hspace*{1em}Emotion risk score = \{R\_der\_only\} -> \{risk\_level\_der\_only\}\\[6pt]
Generate an intervention message in JSON format.
\end{quote}

\section{Judge Prompts}\label{app:judge_prompts}

All Judge evaluations use GPT-5 with temperature fixed at 0.0 to ensure reproducibility.

\subsection*{Tone Classification Prompt ($S_\text{risk}$)}

\begin{quote}
\small
You are an expert judge evaluating the tone of a driver intervention message. Classify the tone as exactly one of: gentle, warning, urgent. Output ONLY the single word with no explanation.\\[6pt]
Message: ``\{message\}''
\end{quote}

\subsection*{Contextual Relevance Prompt ($S_\text{context}$)}

\begin{quote}
\small
You are an expert judge. Given the driver state and message, rate how well the message fits the situation on a scale of 1-5 (1=completely inappropriate, 5=perfectly appropriate). Output ONLY the number.\\[6pt]
Driver: Traffic=\{tcr\_label\}, Vehicle=\{vcr\_label\}, Emotion=\{der\_label\}, Behavior=\{dbr\_label\}, Risk=\{risk\_level\}\\
Message: ``\{message\}''
\end{quote}

\subsection*{Driver Acceptability Prompt ($S_\text{accept}$)}

\begin{quote}
\small
You are an expert judge. Evaluate this driver message on 3 criteria (1-5 each):\\
1. Non-imperative (1=commanding, 5=suggestive)\\
2. Non-threatening (1=threatening, 5=calm)\\
3. Suggestion-style (1=no suggestion, 5=strong suggestion)\\
Output ONLY three numbers separated by commas (e.g. 4,3,5).\\[6pt]
Message: ``\{message\}''
\end{quote}

\end{document}